\documentclass[conference]{ieeeconf}
\IEEEoverridecommandlockouts

\usepackage[backend=biber,
            url=false,
            isbn=false,
            doi=false,
            backref=true,
            style=ieee,
            citestyle=numeric-comp,
            sorting=none,
            block=none]{biblatex}

\usepackage{atbegshi}
\AtBeginDocument{\AtBeginShipoutNext{\AtBeginShipoutDiscard}}

\usepackage{flushend}
\usepackage{multirow}
\usepackage{color}
\usepackage{times}
\usepackage{biblatex}

\usepackage{marginnote}
\usepackage{graphics}
\usepackage[pdftex]{graphicx}
\DeclareGraphicsExtensions{.pdf,.png,.jpg}

\usepackage[rightcaption]{sidecap}
\usepackage{pbox}

\usepackage{mathtools}
\usepackage{amsmath, amssymb, amscd}
\usepackage{ wasysym } 
\usepackage{amsfonts}
\usepackage{mathptmx} 
\usepackage{gensymb} 

\DeclareMathAlphabet{\mathcal}{OMS}{lmsy}{m}{n}
\DeclareSymbolFont{largesymbols}{OMX}{cmex}{m}{n}
\usepackage{textcomp} 

\usepackage[ruled,vlined,linesnumbered]{algorithm2e} 
\usepackage{algorithmicx, algpseudocode} 

\usepackage{array} 
\usepackage{tabularx}
\usepackage{multirow}
\usepackage{multicol}
\usepackage{booktabs}

\usepackage[T1]{fontenc} 
\usepackage[utf8]{inputenc}
\usepackage[english]{babel} 
\usepackage{units}
\usepackage{bm}
\usepackage{times} 
\usepackage{xspace}
\usepackage{flushend}
\usepackage{csquotes}
\usepackage{makeidx}

\let\labelindent\relax
\usepackage{enumitem}

\usepackage{soul} 
\usepackage{subfiles} 

\usepackage[yyyymmdd]{datetime}

\date{\protect\formatdate{1}{1}{2001}}

\usepackage{url}
\makeatletter
\g@addto@macro{\UrlBreaks}{\UrlOrds}
\makeatother
\usepackage{color}
\usepackage[usenames,dvipsnames,table]{xcolor}
\AfterPreamble{\usepackage[pdfborder={0 0 0.5}]{hyperref}
\hypersetup{
    linkcolor=black,
    citecolor=black,
    filecolor=cyan,
    urlcolor=black
}}

\usepackage{soul} 

\newcommand{\todo}[1]{\textcolor{red}{[#1?]}}

\newcommand{\ignore}[1]{}

\newcommand{\seclabel}[1]{\label{sec:#1}}

\usepackage{balance}

\begin{document}
%
\title{SPRK: A Low-Cost Stewart Platform For Motion Study In Surgical Robotics}

\author{%
Vatsal Patel*,
Sanjay Krishnan*,
Aimee Goncalves,
Ken Goldberg
\thanks{*Denotes equal contribution. All authors are affiliated with the AUTOLAB at UC Berkeley: \url{autolab.berkeley.edu}.}}

\maketitle

\begin{abstract}
To simulate body organ motion due to breathing, heart beats, or peristaltic movements,
we designed a low-cost, miniaturized SPRK (Stewart Platform Research Kit) to translate and rotate phantom tissue.
This platform is $20 cm \times 20 cm \times 10 cm$ to fit in the workspace of a da Vinci Research Kit (DVRK) surgical robot 
and costs $\$250$, two orders of magnitude less than a commercial Stewart platform.
The platform has a range of motion of $\pm$ 1.27 cm in translation along $x$, $y$, and $z$ directions and has motion modes for sinusoidal motion and breathing-inspired motion. Modular platform mounts were also designed for pattern cutting and debridement experiments.
The platform's positional controller has a time-constant of 0.2 seconds and the root-mean-square error is 1.22 mm, 1.07 mm, and 0.20 mm in $x$, $y$, and $z$ directions respectively.
All the details, CAD models, and control software for the platform is available at \url{github.com/BerkeleyAutomation/sprk}.
\end{abstract}

\IEEEpeerreviewmaketitle

\section{Introduction}
\seclabel{intro}
Surgical robots, such as Intuitive Surgical's da Vinci, perform more than 500,000 procedures per year \cite{surgical2015annual}. 
The field is rapidly changing with several new innovations in hardware and software~\cite{taylor2016medical,yip2017robot}.
A number of research labs that study extensions in hardware and software for surgical robots use 
animal tissue or synthetic proxies for tissues and anatomy ~\cite{sridhar2017training}.

Training regimens for surgical residents are well-studied (e.g., Robotic Objective Skills Assessment Test ~\cite{siddiqui2014validity}).
Similarly, the medical community has designed a number skill evaluation suites such as the Fundamentals of Laparoscopic Surgery (FLS)~\cite{Ritter2007} and the Fundamental Skills of Robotic Surgery (FSRS)~\cite{frsr-stegemann2013}.
These regimens describe several important tasks that simulate key surgical skills in static environments.
However, these do not simulate body motion due to breathing, heart beats, or peristaltic motion.
Motion due to heart beats is common during bypass procedures ~\cite{buxton2013history}, and peristaltic movements are prevalent in uterine procedures~\cite{kunz2002uterine}.

\begin{figure}[t]
    \centering
    \includegraphics[width=0.49\columnwidth]{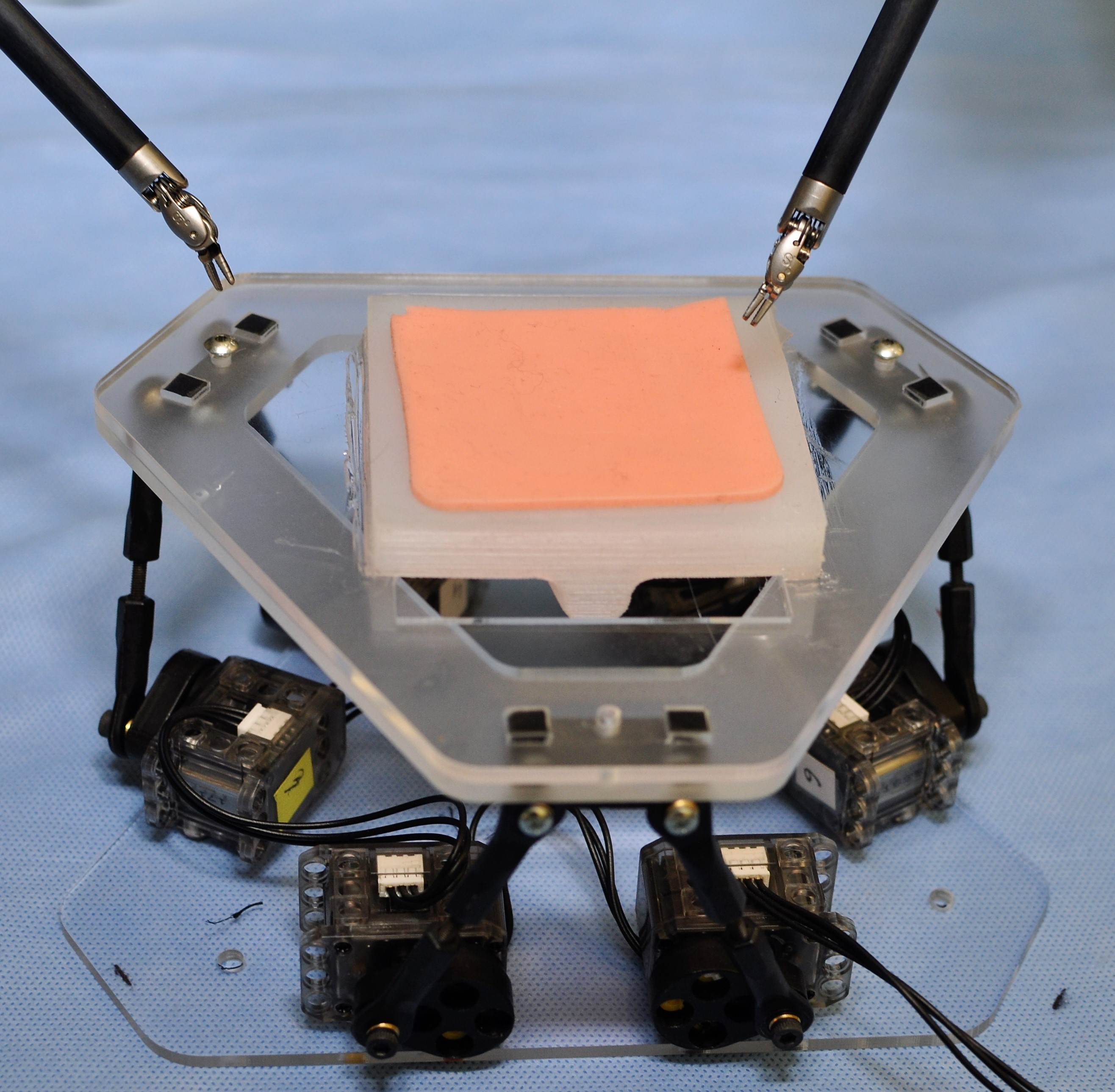}\vspace{0.25em}
     \includegraphics[width=0.49\columnwidth]{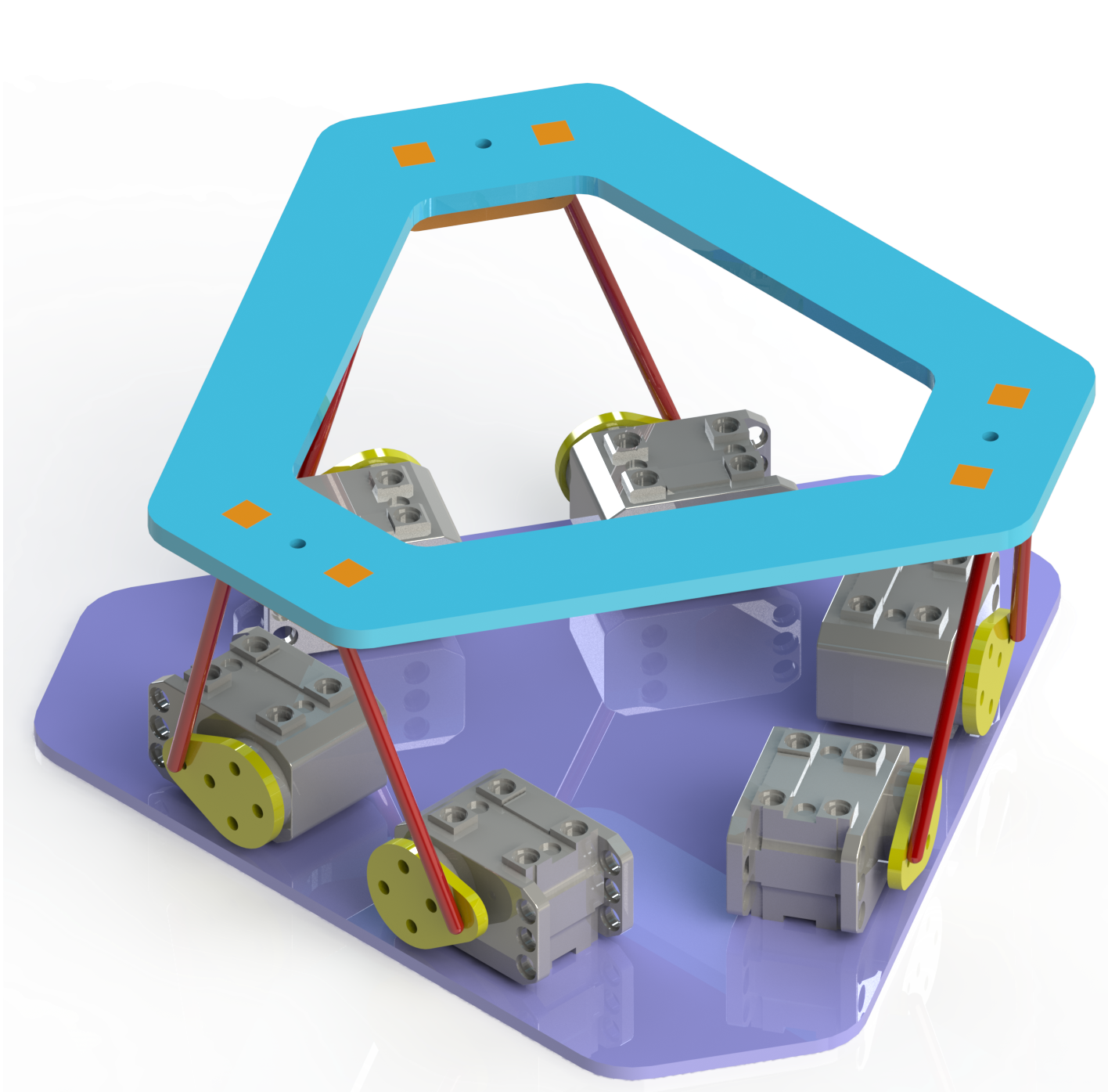}
    \caption{A miniaturized Stewart platform, which is a 6 DoF parallel robot, was developed to simulate anatomical movements during surgery within the da Vinci RSA workspace and used for tasks such as cutting, debridement, and movable cameras.
    Rendered CAD model (right) of the SPRK 2. The bottom structure (purple) and the top platform (cyan) were constructed from 6.35 mm laser cut acrylic sheets, and the servo horns (yellow) and buttress support (orange) were made from 6.35 mm Delrin sheets. The actuators (grey) were Dynamixel XL-320 servos, and the support rods (red) were made using M3 rods and Nylon ball joints. }
    \label{teaser}
\end{figure}

We designed and implemented a low-cost miniaturized Stewart platform 
~\cite{stewart1965platform, goughcontribution1956},
that allows for standard FLS tasks to be implemented, and provides a software interface for inverse kinematics and internal state-estimation for evaluating automation. The platform was used for the study of autonomous subtasks and could be used for training surgeons, although systematic studies were not conducted on the latter.
The platform is built with commercially available components and has a total cost of less than $\$250$.

This paper presents the design of two versions of this platform (SPRK 1 and 2) spanning 18 months of use in the AUTOLAB at UC Berkeley. SPRK 1 met the preliminary requirements of range and frequencies of motion that match breathing or peristaltic movements. SPRK 2 was designed for more accurate and precise motion, especially at larger amplitudes (up to 1 cm) and frequencies (up to 1.5 Hz) of motion that match heartbeat movements. We describe the design of the two versions, the software API, and a characterization of dynamic and kinematic precision.
Our lab has used the SPRK to study autonomous surgical robot in three projects:

\begin{enumerate}
\item \textbf{\emph{Teleoperation: }} In June 2016, co-author and expert cardiac surgeon Dr. W. Douglas Boyd performed two FLS tasks, pattern cutting and peg transfer, on the Stewart platform programmed to move in rhythmic motions. The data collected in this study yielded an interesting insight that the surgeon preferred an \emph{intermittent synchronization} policy, where he synchronized his actions with the minima or maxima of the rhythmic motion (i.e., the lowest velocity time windows) (``Using Intermittent Synchronization to Compensate for Rhythmic Body Motion During Autonomous Surgical Cutting and Debridement'' \cite{stewart2018algorithm}).

\item \textbf{\emph{Surgical Cutting and Debridement: }} We performed autonomous execution on two tasks: (1) we constructed a simplified variant of the FLS cutting task, where we autonomously cut along a line and translated the platform perpendicular to the line at 0.2Hz, and  (2) we consider surgical debridement where foreign inclusions are removed from a tissue phantom that is moving with at 1.25 cm, 0.5 Hz \cite{stewart2018algorithm}.

\item \textbf{\emph{Safe Imitation Learning: }} The platform was used to introduce random physical disturbances in the system during execution of a line tracking task, but not when the system was being trained in order to evaluate the robustness and safety of imitation learning algorithms (``Derivative-Free Failure Avoidance Control for Manipulation using Learned Support Constraints''  \cite{lee2018safeimitation}).
\end{enumerate}

\begin{figure}[t]
    \centering
    \includegraphics[width=\columnwidth]{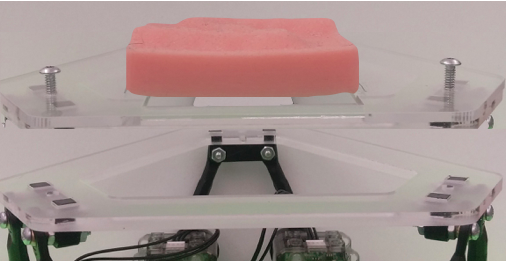}
    \caption{The platform is modular with a removable top mount which allows for mounting tissue phantoms to create configurable FLS task environments. \label{teaser2_modular} }
\end{figure}
\section{Related Work}
\seclabel{rw}

\subsection{Stewart Platforms}
A Stewart platform is a 6 degree of freedom parallel robot that can translate and rotate its platform workspace.  
The device, introduced by Gough \cite{goughcontribution1956} and analyzed by Stewart \cite{stewart1965platform}, was originally designed for tire testing and flight simulation.
Still in use today, pilots are able to train and prepare for various flight scenarios in a cockpit, equipped with screens and aircraft controls, secured to the movable platform. 

The Stewart platform has also been applied to robotic end effectors for precise tool movement \cite{wendlandt1994design, bourges2010assessment,kucuk2016inverse, abbFlexPicker}.
The platform has expanded applications in a variety of fields, including the medical sector. 
The Stewart platform structure has been used for external fixation devices in orthopedics \cite{paley2011history}. 
Girone et al. \cite{girone2001stewart} designed an ankle rehabilitation apparatus that simulates running exercises by utilizing the platform's extensive range of motion and providing resistive force in a virtual reality environment. 
Along the same vein, Yang et al. \cite{yang2015manipulator} developed a handheld "tremor-canceling" tool for surgery by installing the end effector of the tool on the platform of a miniature Stewart platform. 
Wapler et al. \cite{wapler2003stewart} adapted a single, modular platform for use in applications in neuroendoscopy in addition to ophthalmology, spinal surgery, and orthopaedics.


Several variations of parallel manipulators exist on the market, most common of which are hexapods. These are Stewart platforms equipped with six telescoping legs \cite{mikrolarHexapods, symetrieHexapods}. Unlike traditional hexapods, rotary hexapods have a constant leg length and the base attachment points shift to move the top plate \cite{cemecRotapod}. Coulombe and Bonev extended the workspace of rotary hexapods while keeping a small profile by using two concentric rails and double spherical joints \cite{coulombe2013hexapod}. Commercial Stewart platforms can be purchased in a variety of sizes, actuation systems, and system specifications. Industrial platforms from Newport and Physik Instrumente cost over $\$20,000$, plus the controller which costs around $\$10,000$ \cite{newportHexapod, physikHexapod}.

Our design and control of the platform reference \cite{stewartMath} for the kinematics and \cite{stewartInstructable} for manufacturing. The design and kinematics guidelines from these sources were thoroughly optimized for the dimensions of our platform, and for the range and frequency of motion that would be implemented on them.

\subsection{Surgical Robotic Training}
Robotic surgical training techniques and benchmarks are well-studied. 
Motion has been studied in robotic surgery including estimation~\cite{ortmaier2005motion, franke2007improved, ortmaier2003motion, richa2010beating, sloth2012model} and control/compensation ~\cite{cavusoglu2005control, duindam2007geometric, riviere2006robotic, moustris2013active}.
All of this work considers virtual surgical simulators, e.g., Duindam and Sastry \cite{duindam2007geometric}, and proposes a full synchronization approach where the quasi-periodic motion of the anatomy is tracked. Other works, such as Moustris et al. \cite{moustris2013active}, fully synchronize human input on real robot systems with stabilized virtual images, or passively compensate for motion using mounted devices, such as HeartLander \cite{riviere2006robotic}.


\section{Stewart Platform: Technical Description}
This section describes the design of two versions of the platform: SPRK 1 and SPRK 2.


\subsection{Motivation and Design Considerations}

\begin{figure}
    \centering
    \includegraphics[width=0.7\columnwidth]{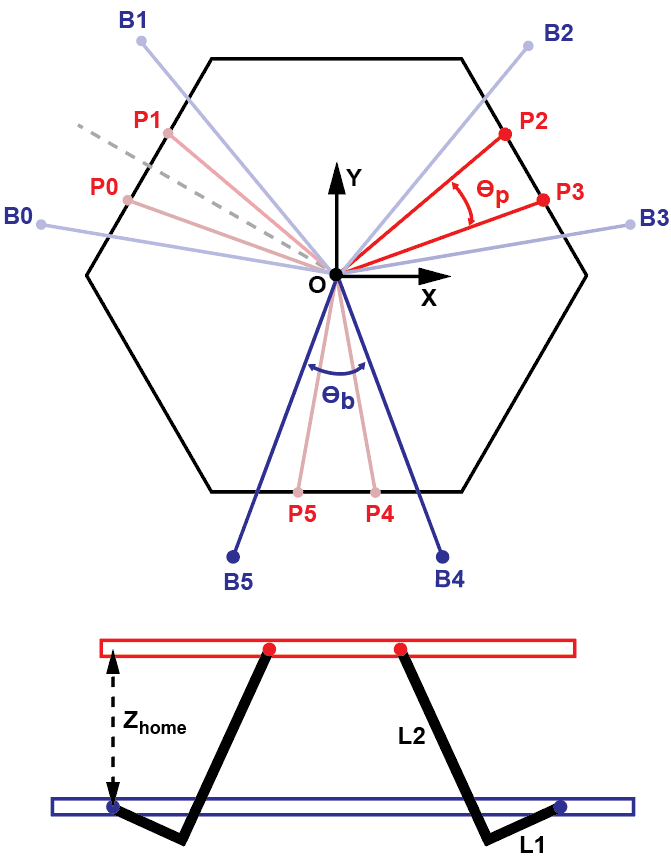}
    \caption{Reference for dimensions of the platform in the top and side view. The attachment points for the support rods on the top (red) plate are indicated as $P_i$, and the attachment points for support rods on the base (blue) actuators are indicated as $B_i$.
    }
    \label{platformDim}
    \renewcommand{\bottomfraction}{0.5}
\end{figure}

The physical and dynamic design of the platform was driven by three specific needs; (1)to have a low-cost 6 DoF platform, (2) to have a range of motion capable of replicating a variety of organ motions, and (3) to fit in the workspace of the da Vinci Research Kit (DVRK). Since internal organs have small ranges of motion \cite{Gagne2014}, we were able to keep the device compact by defining a small desired range of motion.  Additionally, the main constraint regarding physical dimensions was the 21.6 cm between the table of the workspace and the endoscopic stereo camera, driving the device to require a thin vertical profile. The endoscopic camera needed to stay close to the workspace because of its small field of view. Thus, to address (2) and (3), the vertical motion range of the platform is $\pm$ 1.27 cm and the rotational range is $\pm$ $15^{o}$. Fitting under these design constraints, the maximum and minimum vertical height of SPRK 1 is 11.4 cm and 9.5 cm respectively. To further address (3), SPRK 1 had an equally wide base and platform to accommodate our larger surgical simulation tools. Figure \ref{platformDim} shows the dimensions listed in Table \ref{dim_platform}.

Unique to our design was the platform's modular testing surface. Different removable work surfaces can be attached to the main platform by being screwed into a series of preset holes (Figure \ref{teaser2_modular}). An assortment of work surfaces can be created and customized for different surgical tasks and swapped in and out quickly, rather than being constrained to one work surface. It should be pointed out that the device's main top plate was designed with a hollow center. This was done intentionally to prevent any damage to the DVRK's end effectors and to the platform itself, when performing tasks during testing. The platform is built from off-the-shelf parts and easily fabricated components. 

A second version of the Stewart platform (SPRK 2) displayed in Figure \ref{teaser} was made to address some of the limitations of SPRK 1. The acrylic support structures for the servos physical restricted the range of motion of the platform by limiting movement of the rods. The servos on SPRK 1 were also controlled using complicated pulse width modulation technique which caused jittery and inaccurate motion, further amplified by the low resolution and small stall torque of the servos. So, the HI-TEC HS422 servos were replaced with Dynamixel XL-320 servos, and the Teensy micro-controller was replaced with the OpenCM 9.04C controller. The new servos were controlled with digital packets so they offered faster and repeatable motion at higher frequencies. The new micro-controller updated the value on each of the actuators every 5 microseconds for the sinusoidal and breathing motion modes. These servos also had a smaller minimum control angle and larger range of motion which collectively led to smoother motion at higher amplitudes. The higher stall torque of the new motors also improved the durability and motion of the platform by handling accidental knocks, and recovering from high friction positions. SPRK 2 also replaced the servo supports in the front with rivets at the bottom, and used a redesigned servo horn to prevent the rods from coming in contact with any components and limiting the range of motion. The overall kinematics of SPRK 2 were very similar to SPRK 1 because important platform dimensions were not changed in the redesign. The modular work surfaces designed for SPRK 1 could also be used with SPRK 2 because the attachment points layout was retained from SPRK 1. The code on the micro-controller and API scripts used for serial communication were also modified to account for the digital packet communication and the larger platform range of motion.



\begin{table}[]
\centering
\caption{Platform Dimensions used in SPRK 1 and 2, referenced from Figure \ref{platformDim}}
\label{dim_platform}
\begin{tabular}{cll}
\textbf{Variable}                 & \textbf{Value}                       & \multicolumn{1}{c}{\textbf{Description}}        \\ \hline \hline
    $X_{overall}$                   & 18.6 cm                              & Overall length along X-axis \\
    $Y_{overall}$                   & 17.6 cm                              & Overall width along Y-axis \\
    $Z_{overall}$                   & 10.5 cm                              & Overall height along Z-axis \\
    L1                            & 2.0 cm                             & servo arm length      \\
    L2                            & 6.0 cm                             & platform linkage length       \\
    $Z_{home}$                    & 5.1 cm                              & platform height above servo axis (linkages at $90^{o}$)  \\
    $L_{OB}$                      & 8.1 cm                            & length $O$ to platform attachment points $P_{i}$          \\
    $L_{OP}$                      & 8.1 cm                            & length $O$ to servo attachment points $B_{i}$ \\
    $\theta_{b}$                  & $31^\circ$                          & angle between servo axis attachment points \\
    $\theta_{p}$                  & $23.5^\circ$                          & angle between platform attachment points    \\
\end{tabular}
\end{table}

\subsection{Software Interface}

The code with the inverse kinematic calculations, servo communication protocols, and motion mode details is uploaded to the micro-controller. The \texttt{Python} API is used to access these different poses and motion modes. The platform has two preset motion modes available for each degree of freedom. The amplitude and frequency of the oscillations are sent via serial communication to the micro-controller, which updates the actuator positions every 5 microseconds. The actuators on SPRK 2 can reliably perform wave motions at frequencies up to 1.5 Hz. The first available mode is sinusoidal motion, governed by:
\[ y(t) = A\sin(2 \pi \omega t)\]
where $A$ is amplitude and $\omega$ is frequency. This mode can be implemented as translation along an axis, and also as rotation about these axes. The platform can also be commanded to oscillate simultaneously in more than one of those directions.

The second mode simulates breathing-inspired motion, motivated by \cite{stewartBreathe}. Breathing motion is defined as:
%
\[ y(t) = \Big( exp(\sin(\omega t)) - \frac{1}{e} \Big) \frac{2A}{e-\frac{1}{e}}\]
The motion is similar to the sinusoidal wave, but the crests are have a slightly shorter time duration compared to the troughs of the wave. So, the platform spends a longer time around its minima than its maxima. This mode can also be implemented in translation along an axis, or rotation about an axis.
%
%
where $A$ is amplitude and $\omega$ is frequency. 
For rotational motion, the amplitude $A$ is subtracted from the wave so a "breath" oscillates about each axis, rather than the defined amplitude value.




\section{Characterization Experiments}



We characterized accuracy and precision by measuring the correlation between input commands and actual positional movement of the platform. To obtain this relationship, 
readings from encoders on each of the six SPRK 2 actuators were recorded for 150 trials of randomized $x$, $y$, and $z$ translation inputs. A range of platform poses were fed through the inverse kinematics model and the output was compared with the recorded encoder values in order to determine the corresponding actual pose of the platform.

\begin{figure}[h]
\centering
\includegraphics[width=\columnwidth]{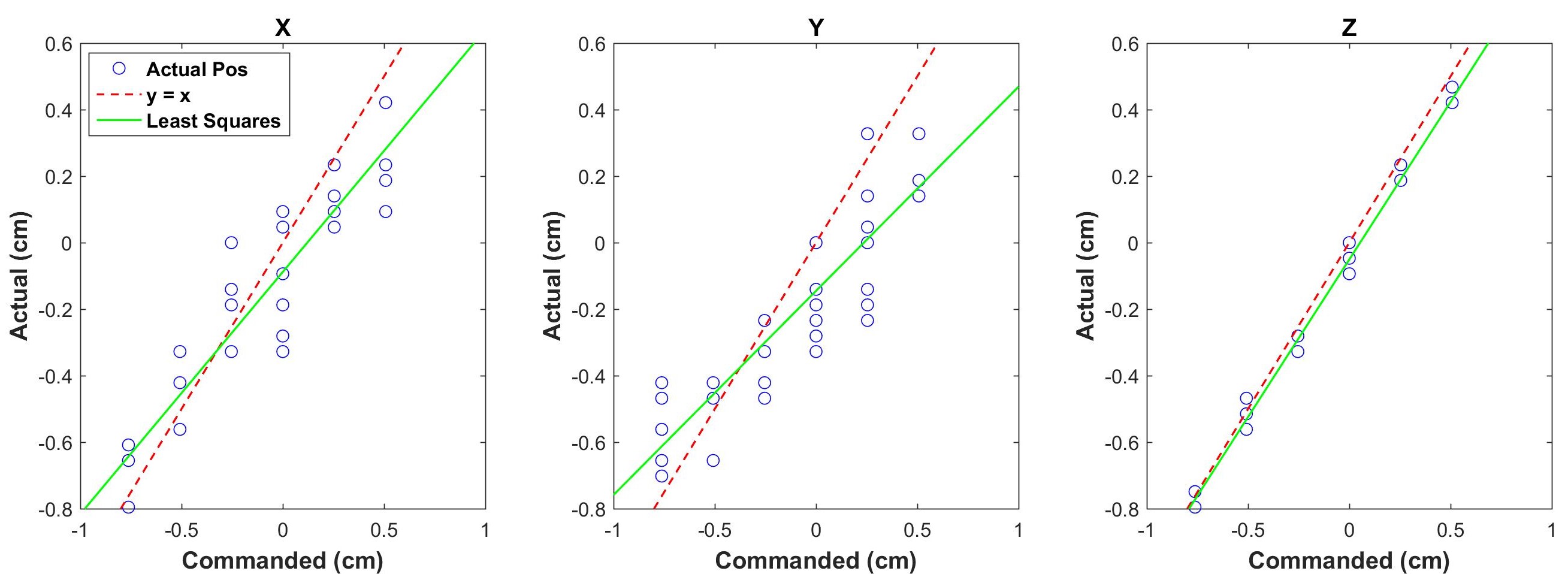}
\caption{Correlation between commanded translation position inputs to and resulting movement from the SPRK 2. Actual platform position was calculated using the readings from the encoders on each of the actuators.
\label{char_stewart} }
\end{figure}

\begin{figure}[h]
\centering
\includegraphics[width=\columnwidth]{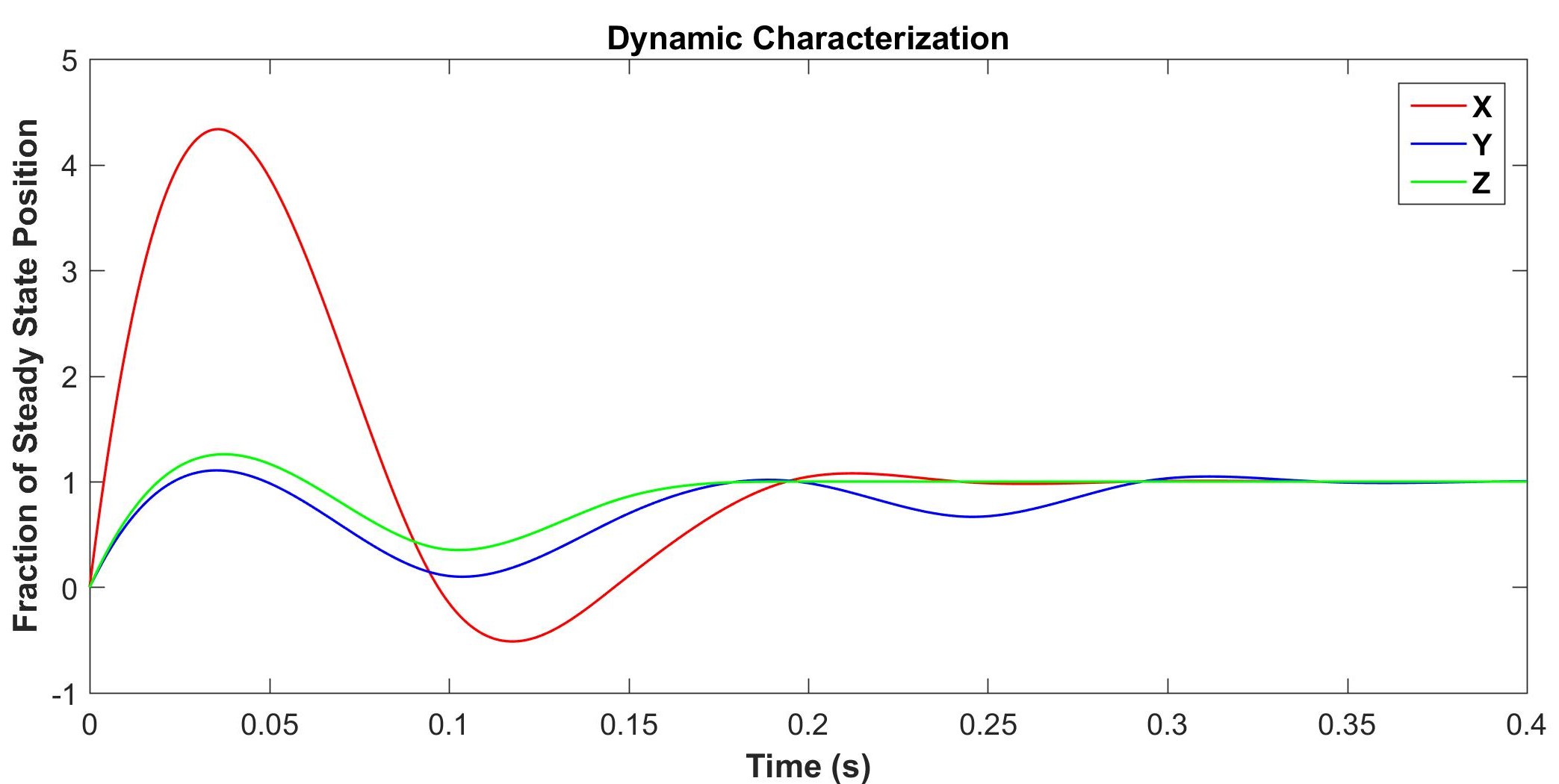}
\caption{Dynamic response of SPRK 2 for a 5 mm step translation input in $x$, $y$, and $z$ directions. The platform reaches the final position within 0.2 to 0.3 seconds. The encoder positions were recorded every 50 ms, and the pose of the platform was calculated using the readings from the encoders on each of the actuators. A cubic spline was fit to the recorded data.
\label{dynchar_stewart} }
\end{figure}

The confidence intervals for the SPRK 1 platform were wide, 
where $R^{2}$ values of $x$, $y$, and $z$ were 0.69, 0.49, and 0.14 respectively. The second version of the platform had significantly improved $R^{2}$ values for $x$, $y$, and $z$ of 0.86, 0.85, and 0.99 respectively (Figure \ref{char_stewart}). These variances are due to several limitations in our system, including kinematic errors in the platform's dynamics, imprecise servos due to step size, and slower encoder readings. Dynamic characterization of the second iteration of the platform confirmed that the settling time in each of the $x$, $y$, and $z$ directions was around 0.2 to 0.3 seconds as shown in (Figure \ref{dynchar_stewart}). The larger overshoot along the $x$ directions compared to that in $y$ and $z$ directions results from more physical interference at the servo attachment points. Moving forward, these static characterizations could allow us to identify portions of the range of motion with reduced accuracy and to correct for systematic problems by relocating attachment points or modifying servo support structures. The settling times and peak overshoots obtained from the dynamic characterization could drive future actuator design considerations.

Our platform's kinematics are uncertain just like that of any anatomical system. Thus, this decoupling of the platform from the controls further evaluates our intermittent synchronization controller's ability to compensate for motion during cutting or debridement in the presence of systematic uncertainty.

\section{Modular Platform Mounts}

\begin{figure}
    \centering
    \includegraphics[width=0.9\columnwidth]{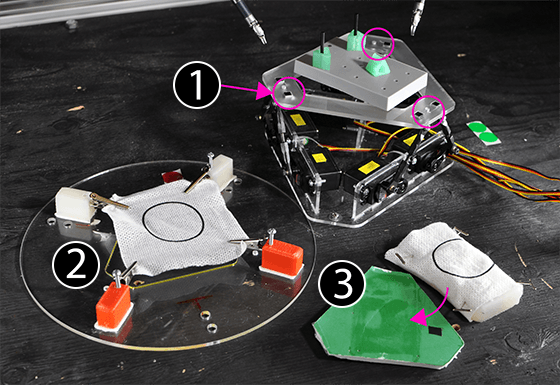}
    \caption{(1) Three attachment points on SPRK 1 (2) Attachment with clips for gauze (3) Attachment for gauze-wrapped silicone phantoms similar to Figure \ref{teaser2_modular}.}
    \label{attach_platform}
    \renewcommand{\bottomfraction}{0.5}
\end{figure}

The modular design of the SPRK 1 and 2 has made them valuable to a wide variety research projects in the lab. These mounts were designed to offer fast experimental reset times while maintaining consistency across experiments. Some of these mounts peg transfer and circle cutting tasks are shown in Figure \ref{attach_platform}. Transparent materials with low reflectivity were chosen for the surfaces to allow for improved visibility and segmentation in the endoscope images. Moreover, safety spaces were opened up in the middle of the platform in order to prevent damage to the surgical instruments and the platform from collisions in failure cases. Pattern cutting and debridement were some of the first experiments conducted using the platform \cite{stewart2018algorithm}. For cutting, the gauze was suspended at the edges with clips, or pinned down on a silicone phantom adhered to the platform mount. For debridement, seeds were placed on a silicone phantom directly attached to the mount. These experiments utilized the rhythmic motion modes of the platform to perform FLS tasks on a physically simulation of anatomical movements. Materials such as gauze, silicone phantoms, and nylon were attached to mounts on the platform's top plate. Prior to these experiments, an expert cardiac surgeon, W. Doug Boyd, performed cutting tasks on the platform under different rhythmic motions and his movements were analyzed to understand the synchronization approaches used. The platform has also been used to induce random physical disturbances in line tracking experiments for developing safe imitation learning policies \cite{lee2018safeimitation}.
\section{Summary and Future Work}\label{ssec:deformation_detection}
Initial results with the SPRK 2 are promising and suggest a number of avenues for future work.


\noindent \textbf{Improved Platform Accuracy and Range of Motion: } In the next design iteration of the platform, we hope to improve the accuracy towards the end of its range of motion in the $x$ and $y$ directions by further reducing points of physical interference and shifting the attachment points of the top plate $60^\circ$ relative to the base to reflect a more traditional Stewart platform design. Better actuators with a faster, more precise response would also help improve the platform's performance near the end of the range.



\noindent \textbf{Simulating Movable Cameras: } 
The clinical Intuitive Surgical da Vinci allows the surgeon to move an endoscope and light during the procedure.
Another intriguing direction for the SPRK is to simulate a movable camera in experimental setups with a stationary camera. In an ideal pinhole camera model the image coordinates $y$ are related to the real-world 3D coordinates $x$ by the camera matrix $C$:
\[
y = C x
\]
We can also consider the inverse problem, namely, for a given camera position what is the corresponding world position:
\[
x = (C^T C)^{\dagger} C^T y 
\]
Suppose, we want to rotate and translate the camera. Let $y' = T y$, where $T \in \mathcal{SE}(3)$. For this $y'$, we want to solve the above inverse problem. Substituting $y = T C x$, we can arrive at the following expression for the corresponding $x'$:
\[
x' = [(C^T C)^{\dagger} C^T T C] x   
\]
\[
x' = [(C^T C)^{\dagger} C^T T C] \approx f_{rigid}(x)   
\]
$f_{rigid}(\cdot)$ is a rigid approximation of the inverse transformation that has to applied to every world frame point to derive the desired transformation of the camera points.
The Stewart platform can implement $f_{rigid}(\cdot)$ for the points in the workspace.

As a use case, consider the following task.
Visual perception of deformable tissue features can require multiple perspectives as they are sensitive to lighting.
We performed a preliminary experiment where we tried to detect ``hyper deformations'' from vision, and this required viewing the material from multiple perspectives. Hyper-deformation is defined as a material that has plastically deformed.
We used a small number of examples of deformed and non-deformed tissue and trained a deep convolutional neural network for the binary classification. For each gauze, we considered eight perspectives. Preliminary experiments yielded about $70\%$ accuracy.


In this paper, we present a modular and low-cost miniaturized Stewart platform to fit in the DVRK workspace, and translate and rotate tissue for autonomous robot surgery experiments and surgical training. The modular surfaces of the platform allows fast and consistent experimental resets for multiple setups including pattern cutting and debridement. A second version (SPRK 2) was designed to improve accuracy and precision at higher frequencies and larger amplitudes. The design of SPRK 1 and 2 supported 18 months of experiments in AUTOlab at UC Berkeley, and the models and code has been made available for use.





{\footnotesize 
\section*{Acknowledgment}
This research was performed at the AUTOLAB at UC Berkeley in
affiliation with the Real-Time Intelligent Secure Execution (RISE) Lab, Berkeley AI Research (BAIR) Lab, and the CITRIS "People and Robots" (CPAR) Initiative: http://robotics.citris-uc.org in affiliation with UC Berkeley's Center for Automation and Learning for Medical Robotics (Cal-MR). This research is supported in part by DHS Award HSHQDC-16-3-00083, NSF CISE Expeditions Award CCF-1139158, by the Scalable Collaborative Human-Robot Learning (SCHooL) Project NSF National Robotics Initiative Award 1734633, and donations from Alibaba, Amazon, Ant Financial, CapitalOne, Ericsson, GE, Google, Huawei, Intel, IBM, Microsoft, Scotiabank, VMware, Siemens, Cisco, Autodesk, Toyota Research, Samsung, Knapp, and Loccioni Inc. We also acknowledge a major equipment grant from Intuitive Surgical and by generous donations from Andy Chou and Susan and Deepak Lim.}

\begin{flushright}
\printbibliography 
\end{flushright}

\end{document}